\pdfoutput=1

\documentclass[11pt]{article}

\usepackage[final]{acl}

\usepackage{times}
\usepackage{latexsym}
\usepackage{graphicx}
\usepackage{hyperref}
\usepackage{tabularx}

\usepackage[T1]{fontenc}

\usepackage[utf8]{inputenc}

\usepackage{microtype}

\usepackage{inconsolata}

\title{Unlocking the Potential of Large Language Models for Clinical Text Anonymization: A Comparative Study}

\author{\bf
David Pissarra,
Isabel Curioso,
João Alveira,
Duarte Pereira,
Bruno Ribeiro,\\\bf
Tomás Souper,
Vasco Gomes,
André V. Carreiro,
Vitor Rolla\\
Fraunhofer AICOS, Portugal\\
\texttt{\{andre.carreiro,vitor.rolla\}@fraunhofer.pt}
}

\begin{document}
\maketitle
\begin{abstract}
Automated clinical text anonymization has the potential to unlock the widespread sharing of textual health data for secondary usage while assuring patient privacy and safety. Despite the proposal of many complex and theoretically successful anonymization solutions in literature, these techniques remain flawed. As such, clinical institutions are still reluctant to apply them for open access to their data. Recent advances in developing Large Language Models (LLMs) pose a promising opportunity to further the field, given their capability to perform various tasks. This paper proposes six new evaluation metrics tailored to the challenges of generative anonymization with LLMs. Moreover, we present a comparative study of LLM-based methods, testing them against two baseline techniques. Our results establish LLM-based models as a reliable alternative to common approaches, paving the way toward trustworthy anonymization of clinical text.
\end{abstract}

{\let\thefootnote\relax\footnote{{Preprint. Under review.}}}

\section{Introduction}

Clinical data contains sensitive information about patients and healthcare professionals. Therefore, information systems must comply with regulations such as the General Data Protection Regulation (GDPR) \cite{gdpr_2018} and the Health Insurance Portability and Accountability Act (HIPAA) \cite{hhs2013hipaa}, which grant data protection rights to the citizens of the European Union (EU) and the United States (US). 


According to the International Organization for Standardization (ISO), data anonymization is “the process by which personal data are irreversibly altered so that a data subject can no longer be identified directly or indirectly, either by the controller or in collaboration with any other party". The anonymization of clinical data ensures that patient privacy is preserved, enabling its sharing. However, in practice, pseudonymization, which involves replacing private identifiers with fake identifiers or pseudonyms, is often more attainable than full anonymization. While pseudonymized data still falls under the scope of regulations like the GDPR, truly anonymized data would not, highlighting the importance of striving for the highest level of data protection possible.
Nevertheless, achieving robust and proper anonymization, especially with unstructured data like clinical notes, is complex. Although many studies \cite{Sweeney1996ReplacingPI,Aramaki2006AutomaticDB,DEHGHAN2015S53,LIU2015S47,YANG2015S30,ocw156,LIU2017S34,friedrich-etal-2019-adversarial} have proposed strategies for the automated anonymization of clinical text, their implementation in real-world contexts is still limited. As a result, access to clinical text data for secondary use remains a barrier to scientific research.

\begin{figure}[!t]
\centering
\includegraphics[width=\columnwidth]{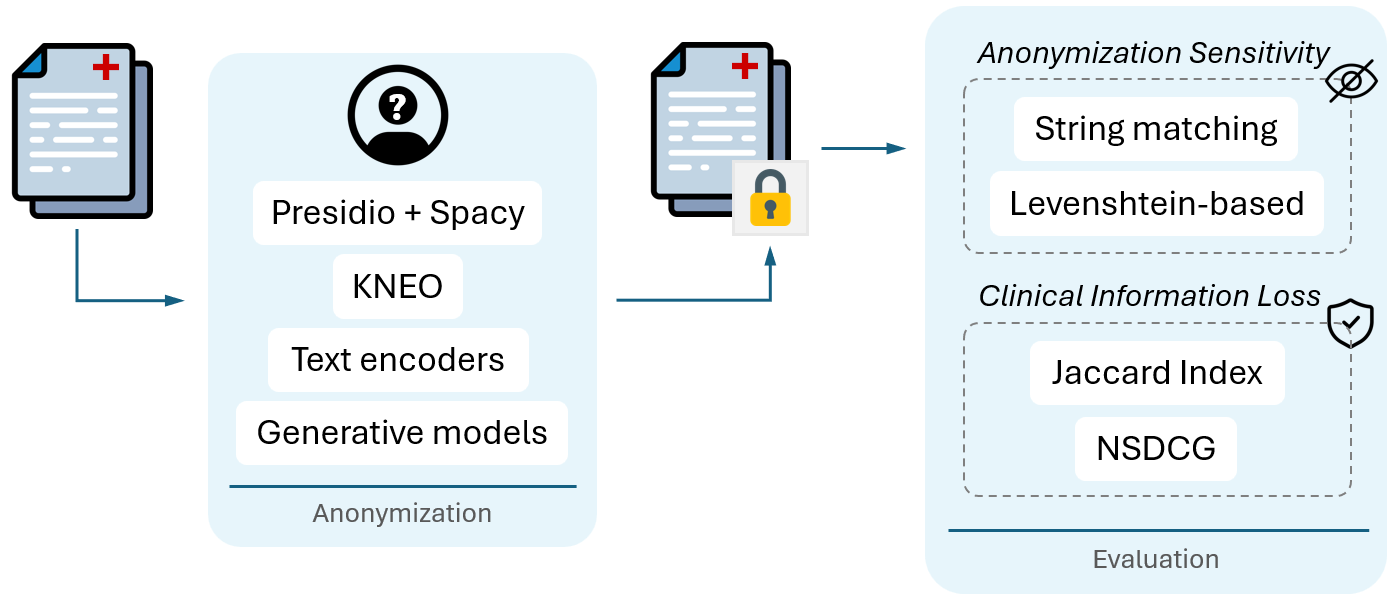}
\caption{Illustration of the followed workflow. Clinical notes can be anonymized through various methods, including LLM-based approaches. A fair evaluation is carried out using novel metrics, compatible with every anonymization strategy.}
\label{fig:pipeline}
\end{figure}

Large Language Models (LLMs) have the potential to be useful tools in anonymizing clinical notes due to their ability to process and interpret vast amounts of unstructured data, produce multilingual text, and leverage extensive general knowledge that may aid in this task \cite{3.5, touvron2023llama}. However, the increasing size of these models raises concerns regarding the inherent sensitivity of this type of data, particularly when using external computing on cloud-based platforms or relying on proprietary models, such as OpenAI's GPT-4 \cite{achiam2023gpt}, which can only be reached through external APIs, potentially compromising confidentiality.


To address this challenge, this work explores the potential of using open-source LLMs that can be locally deployed on cheaper and readily available infrastructure. By running these models locally, healthcare providers can retain full control over their data, significantly mitigating risks associated with external data transfer and storage. Furthermore, local deployment allows for the fine-tuning of models, enhancing the effectiveness of the data anonymization process by adapting the model to the specific types of notes produced by each hospital. This approach ensures the protection of sensitive information and aligns with the growing need for healthcare systems to adopt more secure and regulatory-compliant technologies in handling and analyzing data. To support our approach, this work advances the state-of-the-art by proposing six new evaluation metrics to fairly measure the quality of each model and the clinical information retention in the anonymization process. Figure~\ref{fig:pipeline} illustrates the workflow followed in this paper.

\section{De-Identification Framework and Tools}
\label{chap:incognitus}

The need for effective and reliable clinical text de-identification methods has led to the development of various tools and frameworks. Following this trend, \citet{ribeiro1} developed INCOGNITUS, a toolbox that delivers conventional techniques for automated clinical text de-identification, including a Presidio-based architecture \cite{presidio}, and a de-identification module based on K-Nearest Neighbor Obfuscation (KNEO) \cite{ocaa038}. The goal of this section is to target the background components of the INCOGNITUS framework, which are used as baselines for comparison with the LLM-based methods analyzed in this study.


\subsection{Microsoft Presidio}

Named-Entity Recognition (NER) is a task that aims to identify and classify named entities in text data. In the context of anonymizing clinical notes, NER-based solutions have been historically used to identify and classify sensitive information, such as patients' or doctors' names, IDs, doctor's licenses, dates, phone contacts, emails, professions, hospital names, locations, zip codes, URLs, among other direct or indirect identifiers \cite{DEHGHAN2015S53}.

One practical implementation of this task is Microsoft Presidio \cite{presidio}, an openly available text anonymization tool designed to identify and remove sensitive entities from text data. This tool is composed of two main modules. The first is the analyzer, which identifies sensitive entities based on NER techniques. The second module is the anonymizer, which takes the places associated with those entities and removes or replaces them. INCOGNITUS implements the analyzer module combined with a pre-trained Spacy language model \cite{ines_montani_2020_4091419} and leverages the anonymizer module to produce anonymized text content.

\subsection{KNEO}

While traditional NER-based methods have been reported to achieve high performance in the anonymization task (up to above 90\% recall), search-based methods are always prone to miss certain entities. \citet{ocaa038} alerted to this fact, stating that "as long as current approaches utilize precision and recall to evaluate de-identification algorithms, there will remain a risk of overlooking sensitive information". To address this issue, \citeauthor{ocaa038} proposed an innovative approach that relies on proximity measures between word embeddings to replace every single token of a clinical note with a semantically similar one. This strategy ensures that all sensitive information gets removed, although it raises concerns regarding information loss and readability.

\subsection{Large Language Models}
\label{chap:llms}

LLMs have demonstrated superiority across a wide variety of tasks due to their strong generalization capabilities when trained on significant amounts of data. Their supremacy is attributed to the success of the Transformer architecture \cite{vaswani2017attention} and multiple variants of this architecture have emerged to enhance the performance of LLMs further. As a result, this subset of Deep Learning models is increasingly being adopted in Natural Language Processing (NLP) as a general-purpose language task solver, capable of performing a wide range of language-related tasks, such as text generation, classification, and summarization \cite{llms}.

One notable direction in LLM development is the introduction of encoder-only Transformer models like BERT \cite{devlin2018bert} and decoder-only generative Transformer models such as GPT \cite{radford2018improving}. BERT (Bidirectional Encoder Representations from Transformers) is designed for tasks like natural language understanding and text classification, where bidirectional context is crucial for accurate predictions. On the other hand, GPT (Generative Pre-trained Transformer) focuses on autoregressive text generation and language modelling, demonstrating the capability of LLMs in creative language tasks, such as storytelling and fluent human dialogue.

Within the scope of text anonymization, LLMs have also found significant application \cite{staab2024large}. Text anonymization involves replacing identifiable information in text, such as names, locations, or sensitive details, a task for which textual encoders have been used due to their strong ability to classify and understand sensitive tokens within the text. Generative models, on the other hand, have the ability to recognize sensitive information, such as NER and text-encoder approaches, and also have the potential to recreate content like KNEO, but overcoming its intrinsic limitations like loss of utility and readability. Given LLMs' versatility, we tested both of the aforementioned approaches, encoder-only and decoder-only Transformers on the task of clinical text anonymization.


\subsubsection{Text Encoders}
Text encoders have performed strongly on NER tasks, opening the door for their usage in text anonymization. \citet{devlin2018bert} introduced BERT, an innovative architecture that allows the pre-training of deep bidirectional transformers, and since then, several BERT-variant models have been developed, such as RoBERTa \cite{liu2019roberta} and ALBERT \cite{lan2020albert}. These models can go beyond simple token replacement approaches by leveraging the contextual relevance of sensitive information within the text. Their adaptability allows for successful task-specific fine-tuning, leading to strong performance on problems such as clinical text anonymization \cite{meaney2022comparative}.

\subsubsection{Generative Models}

Deep generative models manifest significant properties in the underlying data-generating process, enabling interpretable representations and controllable generation. The increasing interest in employing generative models in domain-specific tasks, such as within the medical sciences, has propelled this topic into an important area of research. However, deep generative models are not deterministic, and when performing strict tasks such as anonymizing textual content, an intrinsic randomness is associated. The most common generations would involve the removal of sensitive entities, which may be replaced by different types of expressions such as "[REDACTED]" or the symbol "*", the summarization of the overall content with the loss of crucial identifiers, or the removal of small to medium chunks of text.

Third-party LLM APIs (Application Programming Interfaces) like OpenAI's GPT-4 \cite{achiam2023gpt} have exhibited state-of-the-art performance across multiple tasks, especially excelling when provided with prompts for specific use cases. Nevertheless, owing to the success of the open-source community, public foundational generative models \cite{touvron2023llama, jiang2023mistral} have been particularly appealing due to the possibility of adapting them to these domain-specific data resorting to fine-tuning techniques which can crucially enhance their performance.

The key advantage of open-source over proprietary LLMs is the transparency and flexibility it offers to developers. Open-source LLMs provide access to the model's architecture, source code, and training data, allowing for customization of the model to better suit specific goals. More importantly, this also enables local deployment, which mitigates the need to transmit potentially sensitive data, such as medical text containing confidential information, to external servers.

\section{Evaluation Metrics}
\label{chap:eval}

In conventional methods such as NER-based techniques, the computation of commonly used evaluation metrics such as recall, precision, and F1-score is relatively straightforward. Since each token is associated with a label, and classification models output a prediction for each token, one simply needs to compare the true and predicted values to conclude the correctness of each prediction. Nevertheless, using other types of anonymization methods, such as those based on generative models, raises challenges. As these methods output directly anonymized textual content, which may not be written the same way as the input, the link between tokens and labels gets lost. Because the locations of sensitive entities may change from the original to the de-identified version, directly calculating these metrics is no longer possible. Moreover, in generative models not all removed tokens were considered sensitive and thus using the concept of false positive, i.e., a replaced or erased token that did not constitute a sensitive entity, would lead to an unfair judgment of performance.



A potential strategy to identify entities that went unnoticed during anonymization is a total string matching search for the exact content of every sensitive entity in the original content. However, this strategy is limited, as simple alterations to the sensitive entities compromise their detection and, subsequently, the trustworthiness of further calculations based on these matches. To address these issues, we propose four new metrics independent of token-target links that can be used to evaluate any anonymization method fairly. These are built upon the concept of Levenshtein Distance (LD). Furthermore, two of these metrics assess anonymization by focusing more on privacy concerns. It is important to note that text anonymization inevitably entails a trade-off between minimizing privacy risks and retaining data utility. Therefore, we also propose two metrics to evaluate clinical information retention.

\subsection{Levenshtein-based metrics}
The LD quantifies how similar two strings are by measuring the number of deletions, insertions, or substitutions required to transform one string into another. The larger the LD between two strings, the more dissimilar they are~\cite{Levenshtein}.

The Levenshtein Ratio (LRa) is a similarity measure derived from the LD according to the following expression, where $LD(a,b)$ is the LD between two strings $a$ and $b$, and $A$ and $B$ are the respective lengths of each of those strings.

\begin{equation}
    LRa(a,b) = 1 - \frac{LD(a,b)}{\max(A, B)}
\label{eq:lra}
\end{equation}

The LRa provides a value between 0 and 1, where 0 means the two strings are completely dissimilar, and 1 means they are identical.


We first propose two metrics based on the concept of LRa: the Average Levenshtein Index of Dissimilarity (ALID) and the Levenshtein Recall (LR). These aim to capture the effectiveness of anonymization when there is no information about the nature of each token (i.e., whether it constitutes a sensitive entity or not) while tackling the limitations of string matching. The computations of both these metrics are formalized next.
Let us consider a list of length $l$ composed of sensitive entities, $se$, that are in an original clinical note, $ON$, of length $L$. For a certain sensitive entity of this list, $se_i$, we start by computing its length, $e$. We then slide a window of length $e$ across the anonymized note, $AN$, with a step of one character, and compute the LRa between each window and $se_i$. The Levenshtein Similarity Index (LSI) of $se_i$ against $AN$ is given by the following expression, where $w_j$ represents the $j$th window of length $e$ within $AN$.



\begin{equation}
LSI = \max_{j=1}^{L-e} {LRa(se_i,w_j)}
\end{equation}

This measure represents the maximum similarity between $se_i$ and the content of $AN$. Having a list, $S$, of the LSIs measured for each entity contained in $se$, the Average Levenshtein Index of Dissimilarity (ALID) is given as follows, where $\langle S \rangle$ is the mean value of S:

\begin{equation}
ALID = (1 - \langle S \rangle) \times 100
\end{equation}

The second metric, LR, also builds upon the concept of LSI. To calculate LR, each LSI in $S$ is compared to a selected similarity threshold, $th_s$, which was set to 0.85 following experimental findings. Labels with LSI below this threshold are considered de-identified, while entities above this threshold are considered not de-identified. The final value of the metric is given through the traditional computation of recall, dividing the number of de-identified entities by the total number of entities.

\begin{equation}
    {LR@th_{s}} = \frac{\sum_{i=1}^{l}{({S}_{i} < {th}_{s}})}{l} \times 100
\label{eq:levenshtein_recall}
\end{equation}

From a privacy perspective, the evaluation of text anonymization should account for some additional concerns. For instance, not masking a direct identifier, such as a person's full name, is more harmful than not masking a quasi-identifier, e.g., a date. Moreover, direct identification is avoided only if all occurrences of direct identifiers are masked, not just some. With this in mind, and inspired by the work of \citet{Pillan2022}, two additional LR-based metrics are proposed: the Levenshtein Recall for Direct Identifiers (LRDI) and the Levenshtein Recall for Quasi-identifiers (LRQI).

Consider a list of length $l_{di}$ that contains the direct identifiers from $ON$. Let $S_{di}$ be the list of LSIs measured for each direct identifier, also of length $l_{di}$. The LRDI can only take two values: 100 if all occurrences of direct identifiers are considered anonymized or 0 otherwise. This all-or-nothing approach addresses the shortcomings of the standard LR from a privacy perspective.

\begin{equation}
    {LRDI@th_{s}} = \textbf{1} \Bigl( {S}_{di} < {th}_{s} \Bigr) \times 100
\label{eq:levenshtein_recall_di}
\end{equation}

Let $l_{qi}$ be the length of a list that contains the quasi-identifiers from $ON$. Note that $l = l_{di} + l_{qi}$. The LRQI is calculated similarly to the LR but only considering quasi-identifiers.

\begin{equation}
    {LRQI@th_{s}} = \frac{ \sum_{k=1}^{l_{qi}}{({S}_{k} < {th}_{s}}) }{l_{qi}} \times 100
\label{eq:levenshtein_recall_qi}
\end{equation}

In these LD-based metrics, an additional step was implemented in which the LSI is used to find the sentence in $AN$ that is most similar to the sentence in $ON$ where the sensitive entity is located. The metrics were only applied in that sentence, minimizing the likelihood of identifying non-relevant similarities (e.g., the name "Tim" with the first three letters of "time", which has an LRa of 1).


\subsection{Clinical Information Retention metrics}
To assess the impact of anonymization on the preservation of clinical concepts, two new metrics were developed. Their computation leverages an openly available BioBERT model \cite{Lee2020BioBERTAP} pre-trained on a hierarchical classification task of ICD-10 code categories, a coding system designed by the World Health Organization to catalog health conditions \cite{world2004icd}. The outputs of this model before and after the anonymization are compared to estimate lost information.

The first metric is based on the Jaccard Similarity Coefficient (JSC) \cite{jaccard}. The outputs of the BioBERT model are transformed into probabilities through a softmax function, and then a threshold $th_{b}$ is applied, which converts values above it to 1 and those below to 0. Doing so obtains a binary representation of the ICD-10 code categories that the BioBERT model considers present in each note. This study set $th_{b}$ to 0.05 based on experimental findings. Finally, the JSC is computed between the two representations corresponding to the note before and after anonymization. Let $C_{11}$ be the number of classes where both representations have a value of 1 and $C_{01} + C_{10}$ be the number of classes where the representations have different values. The clinical information retention based on the JSC is given as:

\begin{equation}
    {JSC@th_{b}} = \frac{ C_{11} }{ C_{11} + C_{01} + C_{10} } \times 100
\label{eq:cil_jaccard}
\end{equation}

In addition to the JSC, we explored a normalized metric that eliminates the need for setting a threshold. As a result, we propose the Normalized Softmax Discounted Cumulative Gain (NSDCG), based on the widely used NDCG (Normalized Discounted Cumulative Gain) ranking metric \cite{jarvelin2002cumulated}. The main assumption underlying NSDCG is that higher results reflect closer proximity between the original and anonymized logit distributions, indicating a higher degree of similarity between the two distributions and thereby gauging the retained clinical information. The only difference from NDCG is that the discount is obtained from applying the softmax function on the transformer logits, resulting in $sd$ (Equation \ref{eq:softmax_discount}) instead of the common logarithmic discount: $\log(i+1)$. The discount is commonly applied to the gain represented by the relevance score $rel$. Consequently, the $SDCG$ (Softmax Discounted Cumulative Gain, based on the Discounted Cumulative Gain (DCG)) is calculated as follows:

\begin{equation}
SDCG@K = \sum_{i=1}^{K} sd_i \cdot rel_i
\label{eq:dcg}
\end{equation}

As for the discount $sd_i$, let $s$ be the sorted (descending) logits from the original note. The softmax discount, considering the $N$ ICD-10 classes at the i-th position, is given by:

\begin{equation}
sd_i = \frac{e^{s_i}}{\sum_{j=1}^{N} e^{s_j}}
\label{eq:softmax_discount}
\end{equation}

The key advantage of using the softmax discount is that it allows weighting each ICD-10 class logit with more precision, whereas the typical logarithmic discount assigns diminishing importance uniformly across all samples, leading to a weak sensitivity between individual classes. Although this problem could be in some cases mitigated by considering only the top $K$ ranked classes using the $K$ parameter, the variability of the logit outputs can still contribute to this problem persisting with a logarithmic function.

Finally, $rel_i$ represents the relevance of the item at position $i$ in the ranked original logits $z$ (i.e., the logits from the original note ranked according to the anonymized note). This relevance can be achieved as shown in Equation \ref{eq:dcg_relevance}, and it is ensured that $rel_i > 0$.

\begin{equation}
rel_i = e^{z_i}
\label{eq:dcg_relevance}
\end{equation}

As usual, the NSDCG is obtained as the NDCG, dividing the SDCG of the anonymized note by the SDCG of the ideal and original note, being expressed as a percentage value:

\begin{equation}
    NSDCG@K = \frac{SDCG@K}{ISDCG@K} \times 100
\label{eq:ndcg}
\end{equation}

\subsection{Summary}

In summary, six new metrics were proposed for a fair evaluation of clinical de-identification methods. ALID, LR, LRDI, and LRQI leverage the concept of LD and focus on anonymization sensitivity, i.e., assessing whether all sensitive entities have been masked.
On the other hand, JSC and NSDCG measure the retention of clinical information. Table~\ref{table_summary_metrics} provides a brief description of each metric.

\begin{table}[h]
\centering
\small
\begin{tabularx}{\columnwidth}{p{1.4cm}|X}
\hline
Metric & Summary \\
\hline \hline
ALID \includegraphics[height=1.6ex]{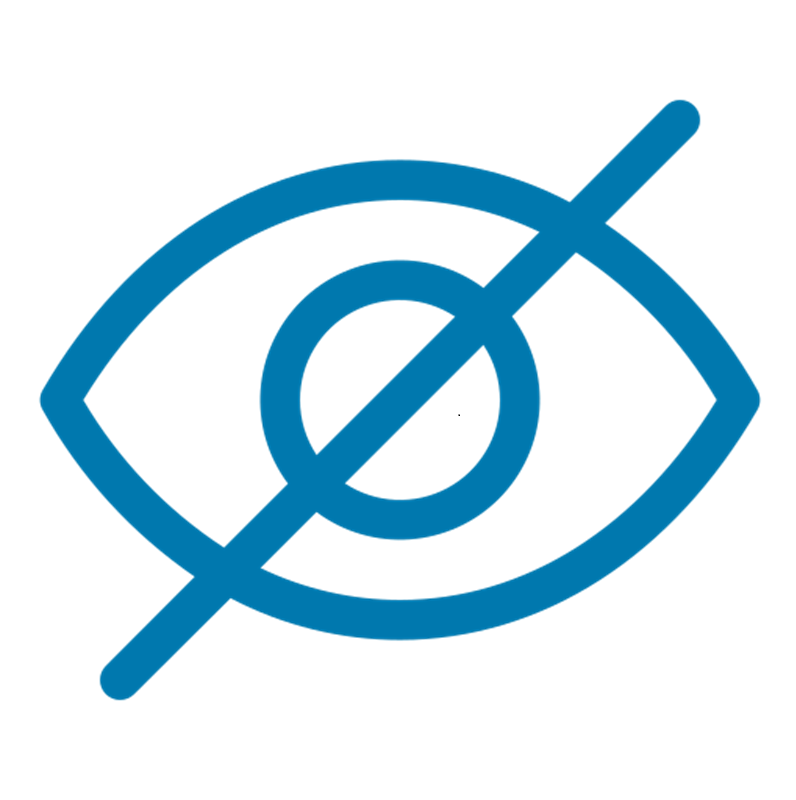} & Complement of the average of the maximum LSI between each sensitive entity and a window of equal length in the AN. \\ \hline
LR \includegraphics[height=1.6ex]{figures/privacy_blue.png} & Proportion of sensitive entities whose maximum LSI with a window of equal length in the AN is below a certain threshold. \\ \hline
LRDI \includegraphics[height=1.6ex]{figures/privacy_blue.png} & LR for direct identifiers. \\ \hline
LRQI \includegraphics[height=1.6ex]{figures/privacy_blue.png} & LR for quasi-identifiers. \\ \hline
JSC \includegraphics[height=1.5ex]{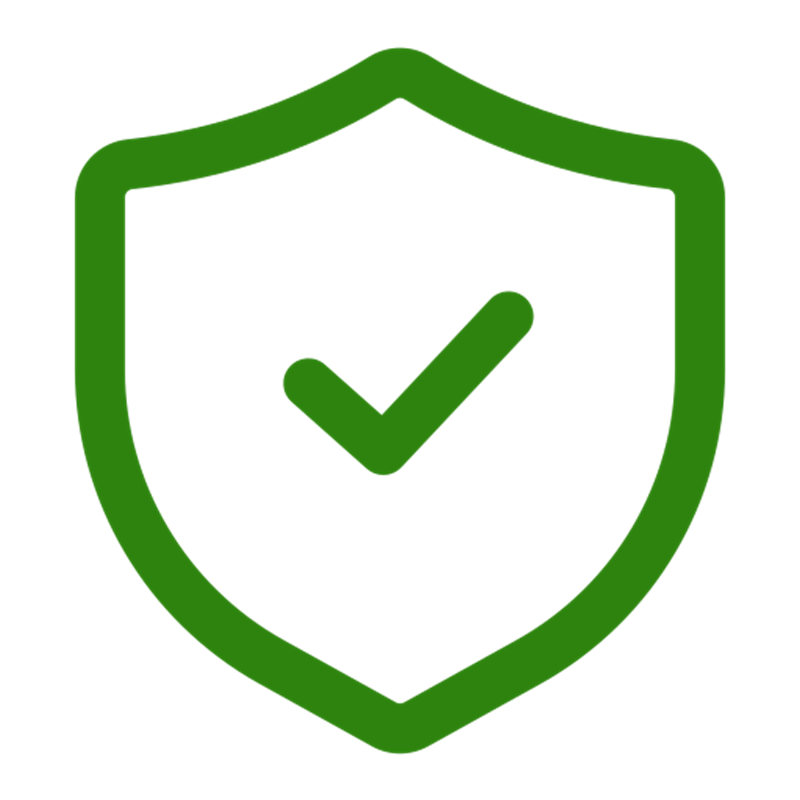} & Jaccard similarity coefficient between the logits from the ON and the AN, after a normalization (softmax) and binarization with a certain threshold. \\ \hline
NSDCG \includegraphics[height=1.5ex]{figures/insurance_green.png} & Normalized Discounted Cumulative Gain with softmax discount. Compares the ranking of the AN's logits against the ON's. \\
\hline
\multicolumn{2}{l}{\hspace{0.1cm}\includegraphics[height=1.6ex]{figures/privacy_blue.png} {\footnotesize Anonymization Sensitivity}} \\
\multicolumn{2}{l}{\hspace{0.1cm}\includegraphics[height=1.5ex]{figures/insurance_green.png} {\footnotesize Clinical Information Retention}} \\
\end{tabularx}

\caption{Summary of the proposed evaluation metrics. The logits mentioned in JSC and NSDCG are from a BioBERT model pre-trained on a hierarchical ICD-10 code categories classification task.}
\label{table_summary_metrics}
\end{table}

\section{Methodology}

The methodology was designed to enable a fair comparison between the performance of different techniques for clinical note anonymization. A total of seven anonymization solutions were compared: two baseline techniques offered by the INCOGNITUS toolbox \cite{ribeiro1}, a fine-tuned BERT model, ClinicalBERT \cite{wang2023optimized}, and four prompt-based methods that leverage Microsoft's Phi-2 and Meta's Llama-3 LLMs \cite{gunasekar2023textbooks, touvron2023llama} (including two zero-shot learning strategies and two fine-tuned models).

\subsection{Dataset}
The experimental dataset includes 66,645 discharge summary notes from the MIMIC III dataset \cite{Johnson2016MIMICIIIAF}. From these, 50\% were used for model training, 20\% for validation, and 30\% for testing. The MIMIC III dataset includes different types of clinical notes (e.g., Nursing, Radiology, and ECG) in different proportions. Therefore, when splitting the data, we ensured that the original distribution remained the same for each subset. Since this dataset was originally anonymized, fake sensitive entities were introduced by employing the Faker library for Python \cite{faker}. This was performed according to the categories of the anonymization tags available in the dataset (e.g. names, phone numbers, emails). The pre-trained LLMs and Presidio only use the test set for inference. All other models were fine-tuned on the training set and validated on the validation set, prior to inference on the test set.

\subsection{Baseline Techniques}


As baselines, we use two techniques from the INCOGNITUS toolbox. The first combines Presidio's analyzer module with a spacy language model, which was pre-trained on the NER task against the OntoNotes 5 dataset \cite{OntoNotes5}. In particular, we used the \textit{en\_core\_web\_trf} English transformer pipeline from spacy, which utilizes a RoBERTa-based model to perform this task. The second baseline method applies a KNEO approach, leveraging a Word2Vec embeddings model. The original anonymized version of the MIMIC III notes was used to ensure that these embeddings did not contain any sensitive information.

\subsection{LLM-Based Anonymization}

For LLM-based methods, ClinicalBERT was fine-tuned on the NER task. To guarantee that no information got lost, some sentences were split into smaller chunks to fit within the maximum context length of the model. Regarding prompt-based models, a system prompt was designed to guide the model in performing anonymization tasks effectively. In this approach, the system prompt serves as an initial instruction or context provided to the generative model, for instance specifying examples of sensitive entities, aiding the model in understanding how it should process and transform the input data. While the system prompt can be very useful in zero-shot inference, where the model has not been specifically trained on anonymization tasks, it also provides a foundation for further fine-tuning. For that reason, we fine-tuned both Phi-2 and Llama-3-8B using the same system prompt to enhance their ability to anonymize clinical text accurately and retain crucial clinical context.

For all trained LLMs, fine-tuning took place on a single 40GB A100 GPU. However, for the largest model (i.e., Llama-3-8B) QLoRA \cite{dettmers2024qlora} was employed to minimize VRAM usage and fit within the GPU's capacity limit.

\subsection{Evaluation}
Each technique was tested on 19,994 notes randomly selected from the dataset. The anonymized versions of these clinical notes were taken along with the original notes to compute the metrics introduced in Section~\ref{chap:eval}: ALID, LR, LRDI, LRQI, JSC and NSDCG. Additionally, a measure of String Matching-based Recall (SMR) was also included.
For the calculation of the privacy risk metrics, i.e. LRDI and LRQI, the categories of the MIMIC III anonymization tags were split as follows: NAME, CONTACT\_NUMBER, ID, and EMAIL were considered direct identifiers, while LOCATION, DATE, URL, AGE\_ABOVE\_89, INSTITUTION, and HOLIDAY were considered quasi-identifiers. A conservative approach was taken to perform this division, i.e., if there is a slight possibility that a category contains personal identifying information, then it is regarded as a direct identifier.

None of these metrics requires a connection between the tokens of the anonymized notes and the sensitive information tags, which makes them compatible with every anonymization method tested. This evaluation strategy allowed for a fair comparison between fundamentally distinct methods, clarifying where LLM-based techniques position in the clinical text anonymization task.

\section{Results and Discussion}

\begin{figure*}[!h]
    \centering
    \includegraphics[width=0.97\textwidth]{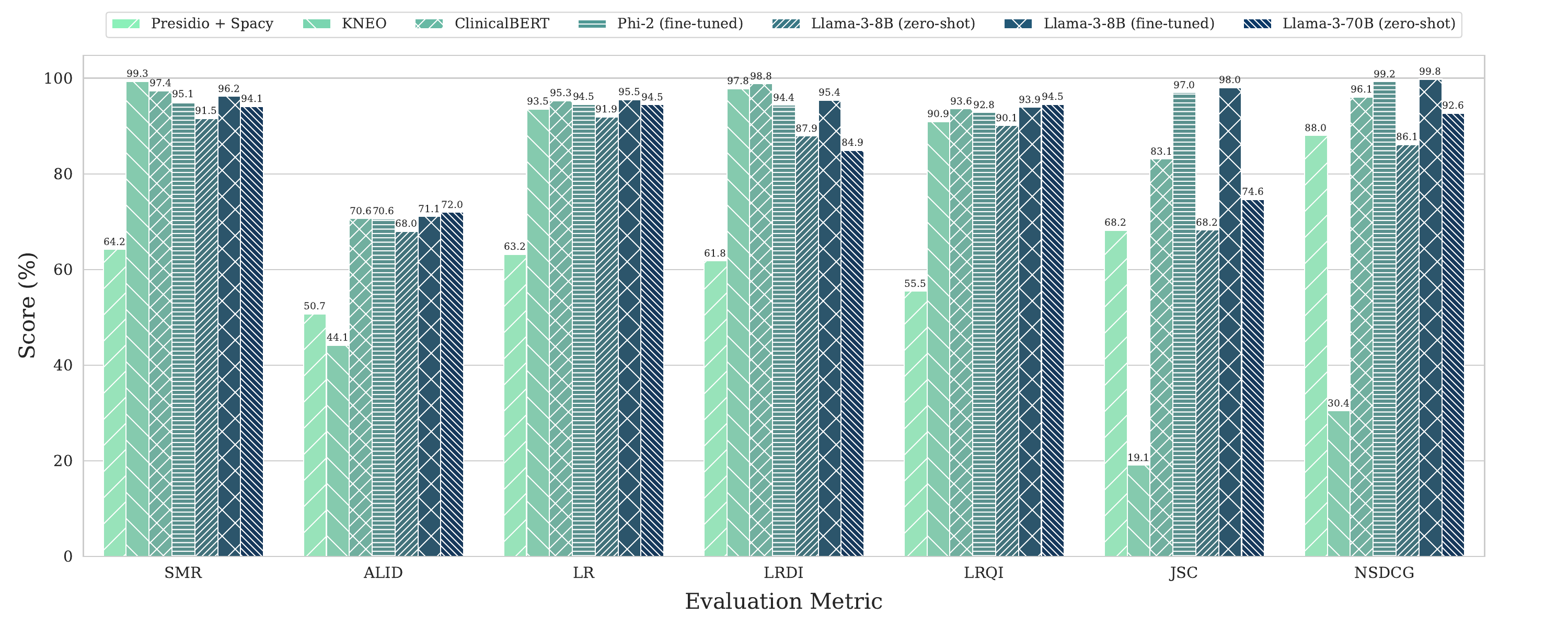}
    \caption{Performance results attained through each anonymization technique tested for seven different evaluation metrics. The results are presented as the average of the metrics measured across 19,994 notes used for testing.}
    \label{fig:anonymisation_results}
\end{figure*}

Figure \ref{fig:anonymisation_results} presents the performances of the different strategies, as given by the average of each evaluation metric measured across all the test notes.

Firstly, we focus on metrics of anonymization sensitivity, particularly SMR and LR. The overall performances measured by both metrics are consistent with one another, with the exception of KNEO which is the best-performing strategy only according to SMR. This result was expected given that this anonymization method replaces every single token, ensuring that no sensitive entity remains unaltered and thus reporting better performances when evaluated by a metric that looks for total string matching. Since LR is not as sensitive to slight alterations in the spelling of entities, one can infer that this metric considered that some changes carried out by KNEO were insufficient to achieve anonymization. This hypothesis is corroborated by the fact that KNEO obtained the lowest ALID, which indicates that the replacements made are less substantial compared to other anonymization methods.

Another noteworthy result is that no anonymization method was able to achieve 100\% in any recall measure, i.e. SMR, LR, LRDI, and LRQI. Further inspection of the data showed that some sensitive entities consist of a pair of letters (name initials), which can easily appear in the middle of non-sensitive words. In addition, there are also some inconsistencies in the labeling of the MIMIC~III dataset, e.g., isolated numbers that are labeled as dates when they merely refer to quantities. These occurrences were misclassified as errors, which explains the absence of perfect recalls. Moreover, although LD-based metrics may identify incomplete de-identification occurrences, they can also be misleading when certain words are similar despite being unrelated (e.g., the name "Tim" and the first three letters of the word "time" produce an LRa of 1). Note that these situations can affect every method, and therefore the comparisons between different techniques are valid nevertheless. As for ALID, its results would never reach 100\% because, even in perfect anonymization, there is always a residual similarity between an entity and any other token in the note. As a result, the observed consistency between the values of the leading approaches might be indicative of a standard plateau of LSI, impacted by some isolated abnormal cases, such as the "Tim-time" pair discussed before.

ClinicalBERT shows the overall best performance on the anonymization sensitivity metrics, while Presidio has the lowest results in 4 out of 5 measurements. Even though Presidio is specifically tailored for text de-identification, it still lacks critical clinical concepts to achieve better performance on clinical-related de-identification. When it comes to information retention, the fine-tuned generative models reported the highest values in both JSC and NSDCG, followed by ClinicalBERT. The most prominent result is the significant difference between the low values achieved by KNEO and those of the remaining methods. This exposes the unfortunate yet somewhat expected outcome related to the significant loss of information associated with the application of KNEO, reflected both in a lower JSC ($19.1\%$ compared to an average of $72.6\%$), and in a lower NSDCG ($30.4\%$ compared to an average of $84.6\%$). Considering that the ICD-10 classification model used was trained to classify code categories and not specific codes, this is an even more concerning outcome, which shows that the KNEO strategy needs improvement before being considered a reliable method. Apart from KNEO, our second baseline, Presidio, also underperformed on these two metrics. As a result, in terms of clinical information retention, its performance can be compared to the performance of zero-shot generative models, which also were not specifically pre-trained on the clinical domain.

Looking specifically at the generative methods, all of them were consistent in terms of recall. However, one can notice that an increase in the number of parameters of the model (e.g., Phi-2 has approximately 2 billion, Llama-3-8B has 8 billion parameters, and so on) has a slight positive impact across all metrics. The number of parameters positively correlates with the amount of information distilled into the model, which can enable the model to better generalize across multiple tasks.

Another important point to note is that the tested generative models improve when fine-tuned, while zero-shot models struggle to identify the structure of clinical notes. Although recall metrics are not heavily compromised, zero-shot models often end up anonymizing entities that should not be omitted. For that reason, the precision and clinical information retention of the model are weaker. On the other hand, fine-tuned models have a better understanding of the structure of the clinical text and are able to retain crucial information while anonymizing sensitive entities. Therefore, while increasing the number of parameters improves overall performance, fine-tuning is essential for maximizing the model's precision and its ability to not lose important clinical information. As an example, even the smallest fine-tuned model, Phi-2, was able to beat the largest zero-shot model, Llama-3-70B, on both clinical information retention metrics, while keeping competitive recall results.

\section{Conclusions}

This work presents a comprehensive comparative study between traditional methods for the automated anonymization of clinical text and new techniques that leverage the power of LLMs. Two different approaches from the INCOGNITUS anonymization toolbox and five methods based on LLMs were tested across seven different performance metrics, including six newly proposed metrics designed to tackle the challenges inherent to generative methods.
The results introduce anonymization techniques based on LLMs as a promising alternative to the current methods, representing a step forward toward unlocking the true potential of clinical text data for secondary usage.

\section{Limitations}

Regarding the proposed evaluation metrics, we believe there are opportunities for improvement in future work. Despite having advantages compared to total string matching, a limitation of LD-based metrics is the identification of strong similarities between entities and unrelated text spans, e.g., "Tim" and "time". This may lead to an underestimation of the performance, which, from a cautious and privacy risk perspective, is still preferred over the overestimation that total string matching might entail. Furthermore, the LRQI evaluates each entity separately, thus disregarding the combined effect of quasi-identifiers, which increases the privacy risk. Also, the binarization step performed in the JSC calculation renders this metric insensitive to differences in the values of each class between the original and anonymized logit distributions, as it only compares the presence/absence of classes. Finally, LR, LRDI, LRQI, and JSC are all dependent on thresholds, which may require adjustments for each case study. Determining the optimal threshold poses a challenge for effective model evaluation and may impact the consistency across different datasets and contexts.

Another significant aspect to note is that using a BioBERT to compare logit distributions within the scope of information retention can sometimes be faulty in the presence of clinical notes with a higher degree of anonymization. The reason for this is that the text classifier was not specifically fine-tuned on anonymized text, and even slight deviations from the typical structure of a clinical note can result in flawed logit outputs, affecting the precision of the information retention metrics. Additionally, the information retention measured by these metrics is based on a BioBERT model pre-trained on ICD-10 classification, which might not be the most reliable ground truth. This reliance can introduce biases and limit the generalizability of the results. Future research should consider developing more robust and contextually relevant ground truth models for better evaluation accuracy.

In conclusion, while the proposed evaluation metrics represent a significant step forward in assessing the performance of LLM-based anonymization techniques, addressing these limitations is crucial for further refining and enhancing their reliability and applicability.

\bibliography{anthology,custom}


\end{document}